\pdfoutput=1
\documentclass[11pt]{article}

\usepackage{acl}

\usepackage{times}
\usepackage{latexsym}

\usepackage[T1]{fontenc}
\usepackage[utf8]{inputenc}

\usepackage{microtype}

\usepackage{inconsolata}

\usepackage{graphicx}

\usepackage[skip=3pt]{caption}
\usepackage[font=10pt,skip=3pt]{caption}
\usepackage{soul}
\usepackage{url}
\usepackage{amsmath}
\usepackage{amsthm}
\usepackage{booktabs}
\usepackage{algorithm}
\usepackage{algorithmic}
\usepackage[switch]{lineno}
\usepackage{multirow}

\usepackage{float}

\usepackage{tcolorbox} % For creating colored boxes
\usepackage{enumitem}   % For customizing itemize environment
\usepackage{pifont}     % For special symbols (ding)
\usepackage{setspace}   % For custom spacing between items

\usepackage{todonotes}

\title{ClaimVer: Explainable Claim-Level Verification and Evidence Attribution of Text Through Knowledge Graphs}

\author{
  {\bf Preetam Prabhu Srikar Dammu}\textsuperscript{1}, 
  {\bf Himanshu Naidu}\textsuperscript{1}, 
  {\bf Mouly Dewan} \textsuperscript{1}, \\
  {\bf YoungMin Kim} \textsuperscript{1},
  {\bf Tanya Roosta}\textsuperscript{2,4,\thanks{Work does not relate to position at Amazon.}}, 
  {\bf Aman Chadha}\textsuperscript{3,4,\footnotemark[1]}, 
  {\bf Chirag Shah}\textsuperscript{1} \\
  \textsuperscript{1}University of Washington \hspace{1em} \textsuperscript{2}UC Berkeley  \hspace{1em}%\\ 
  \textsuperscript{3}Stanford University \hspace{1em} \textsuperscript{4}Amazon GenAI \\
}

\begin{document}
\maketitle

\begin{abstract}
    In the midst of widespread misinformation and disinformation through social media and the proliferation of AI-generated texts, it has become increasingly difficult for people to validate and trust information they encounter. Many fact-checking approaches and tools have been developed, but they often lack appropriate explainability or granularity to be useful in various contexts. A text validation method that is easy to use, accessible, and can perform fine-grained evidence attribution has become crucial. More importantly, building user trust in such a method requires presenting the rationale behind each prediction, as research shows this significantly influences people's belief in automated systems. Localizing and bringing users' attention to the specific problematic content is also paramount, instead of providing simple blanket labels. In this paper, we present \textit{ClaimVer, a human-centric framework} tailored to meet users' informational and verification needs by generating rich annotations and thereby reducing cognitive load.  Designed to deliver comprehensive evaluations of texts, it highlights each claim, verifies it against a trusted knowledge graph (KG), presents the evidence, and provides succinct, clear explanations for each claim prediction. Finally, our framework introduces an attribution score, enhancing applicability across a wide range of downstream tasks.
\end{abstract}

\begin{figure}[t]
    \centering
    \includegraphics[width=0.9\linewidth]{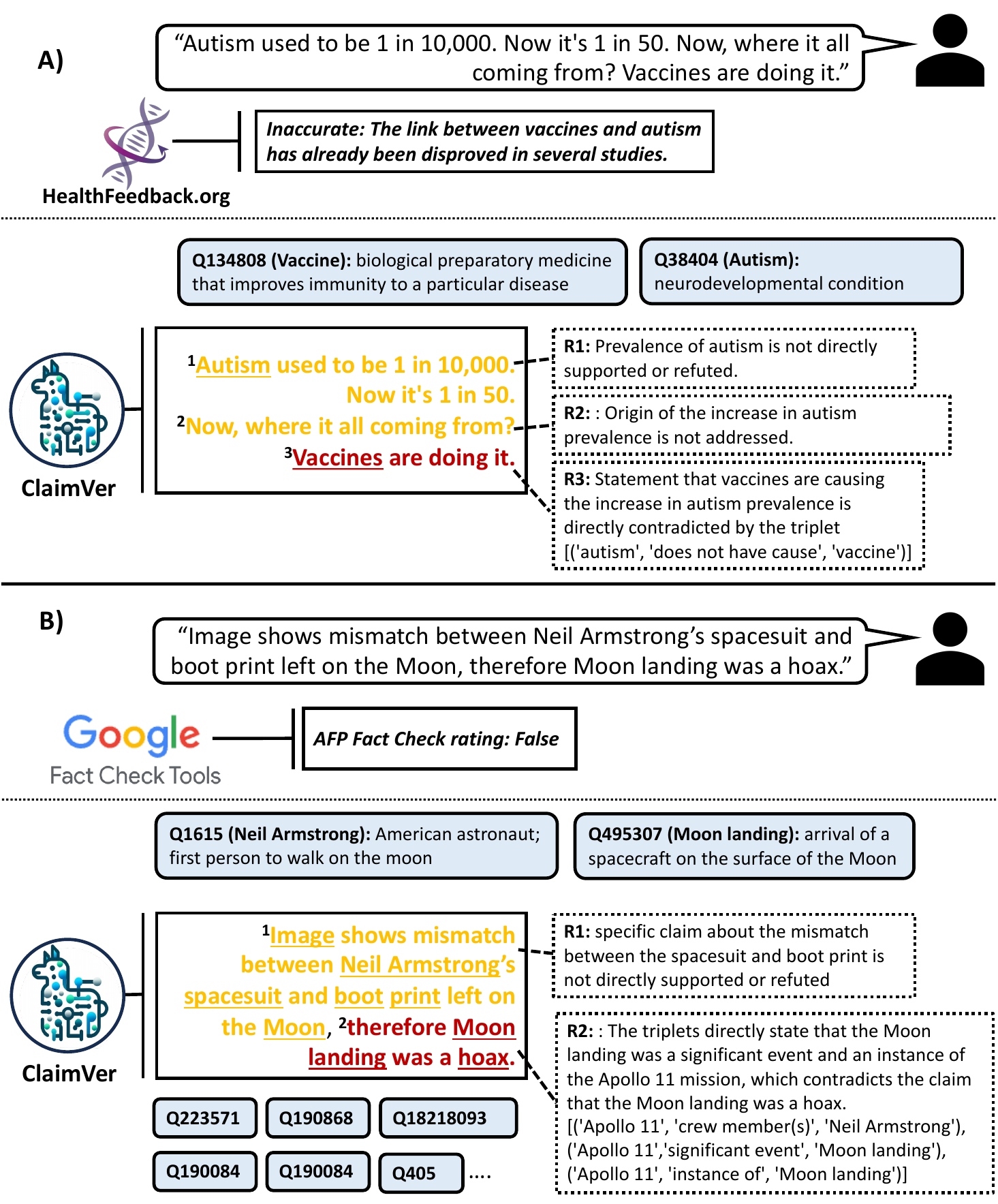}
    \caption{Demonstration of ClaimVer for claim verification and evidence attribution. (A) Text labeled as \textit{Inaccurate} by HealthFeedback and ClaimVer's predictions, rationale, and evidence. (B) Text labeled as \textit{False} by Google Fact Check Tools and ClaimVer's outputs. Predictions are color-coded (amber: extrapolatory, red: contradictory); $R_i$: rationale; related wiki entities are displayed in boxes.}
    \label{fig:teaser}
    \vspace{-5mm}
\end{figure}

\section{Introduction}

Misinformation and disinformation are longstanding issues, but the proliferation of AI tools that can generate information on demand has amplified these issues. Tools for fact-checking are not keeping pace with sophisticated text generation techniques. Even when they are effective, they lack appropriate explainability and granularity to be useful to users. Studies have shown that explanations are crucial for users to build trust in AI systems \cite{rechkemmer2022confidence,weitz2019you,shin2021effects}. Therefore, there is a need for a novel human-centric approach to text verification that offers usable and sufficiently granular explanations to inform and educate the user.

Most fact-checkers, including widely used ones in deployment, issue blanket predictions that can lead to user misunderstandings. For instance, in Figure \ref{fig:teaser} (A), we observe that HealthFeedback\footnote{\href{https://healthfeedback.org/claimreview/vaccines-do-not-cause-autism-diabetes-or-cancer-contrary-to-shiv-chopras-claim/}{https://healthfeedback.org/}}, a fact-checker for medical text, indicates that a misleading statement about the increase in Autism is inaccurate. However, there are multiple claims made in that text, which are not addressed by this tool. In fact, research does show that Autism cases have increased, but this is mostly attributed to increased testing \cite{russell2015changes}. Our method accurately breaks down the text into multiple claims and shows that the specific claim that vaccines are causing autism is indeed incorrect, attributing it to a fact from the Wikidata \cite{vrandevcic2014wikidata}. It also provides a clear rationale as to why the first two claims cannot be determined, as there's no conclusive evidence present in the KG. Such granular predictions, supported by justifications, significantly improve user confidence \cite{rechkemmer2022confidence,weitz2019you,shin2021effects}.

Similarly, in Figure \ref{fig:teaser} (B), we notice that Google Fact Check Tools\footnote{\href{https://toolbox.google.com/factcheck/explorer/search/moon\%20landing\%20hoax;hl=en?authuser=0}{https://toolbox.google.com/factcheck/explorer}} provides a blanket label for an utterance denying the moon landing. In contrast, ClaimVer identifies the exact text span that can be conclusively proven incorrect and proceeds to provide specific information about the Apollo 11 mission and its crew members to refute the claim. All verified entities present in the text, along with their Wiki IDs and descriptions, are displayed for user reference.

Prior research \cite{rashkin2023measuring,yue2023automatic,thorne-etal-2019-fever2,aly2021feverous} typically validates text at the paragraph or sentence level without adequately enhancing user awareness by supplying key details such as rationale, match scores, or evidence. A KG-based approach allows for finer granularity, aiding in pinpointing specific inaccuracies like hallucinations in LLM-generated text or false claims in misleading text. Furthermore, if needed, broader-level metrics can be extracted from this detailed attribution.

The assumption of one-to-one mapping between input and reference texts, prevalent in previous methods \cite{rashkin2023measuring,yue2023automatic,thorne-etal-2019-fever2,aly2021feverous}, does not hold if the given text consists of claims that can be mapped to more than one source. In contrast, utilizing a KG, which represents a consolidated body of knowledge, results in a more comprehensive evaluation. While most previous methods may not support scenarios with information spread across various references, querying a KG can yield triplets originally sourced from multiple documents. Additionally, procuring the specific spans of text required to evaluate claims from large text sources that may span several pages presents many challenges. In contrast, a KG captures only the most important relationships as nodes and links, providing a more efficient way to evaluate the claims.

% Prior methods that depend on document indices or vector databases are not easy to maintain or audit. In contrast, existing trusted KGs that are constructed through human curation provide an effective and human-centered approach for evaluating text at scale. Therefore, we leverage KGs to build a framework that realizes our goal of performing fine-grained text verification and evidence attribution. Our framework also generates insights that boost user awareness, thereby fostering increased trust in automated systems.

\section{Related Work}

Research on validating text has been ongoing for the past decade, while the concept of evidence attribution has gained increased attention in recent years, following the advent of generative models. 

% Our method integrates fact verification and evidence attribution, we discuss recent advancements in both domains in this section.
Our method integrates fact verification and evidence attribution. In this section, we discuss recent advancements in both domains.

\subsection{Fact Verification}

Fact verification is a task that is closely related to natural language inference (NLI) \cite{conneau2017supervised,schick2020exploiting}, in which given a premise, the task is to verify whether a hypothesis is an entailment, contradiction, or neutral. Similarly, in fact verification, the task is to check if a given text can be supported, refuted, or indeterminable, given a reference text. Recent studies in this domain show that LLMs can achieve high performance, and can be considerably reliable for verification tasks, even though they are prone to hallucations \cite{guan2023language}.

In \citet{lee2020language}, the authors show that the inherent knowledge of LLMs could be used to perform fact verification. Other works \cite{yao2022react,jiang2023active} have shown that using external knowledge is helpful for many reasoning-intensive tasks, and report enhanced performance on HotPotQA \cite{yang2018hotpotqa} and FEVER \cite{thorne2018fever}. A wide variety of studies have established that LLMs are suitable for fact verification. For example, \cite{dong2021structural} enhanced accuracy of table-based fact verification by incorporating column-level cell rank information into pre-training. In FactScore, authors \cite{min2023factscore}, introduce a new evaluation that breaks a long-form text generated by large language models (LMs) into individual atomic facts and calculates the proportion of these atomic facts that are substantiated by a credible knowledge base.

\subsection{Evidence Attribution}

The distinction between evidence attribution and fact verification lies in the emphasis on identifying a source that can be attributed to the information. This task is becoming increasingly important, as generative models produce useful and impressive outputs, but without a frame of reference to validate them. In \cite{rashkin2023measuring}, the authors present a framework named \textit{AIS} (Attributable to Identified Sources) that specifies annotation guidelines and underlines the importance of attributing text to an external, verifiable, and independent source. \cite{yue2023automatic} demonstrate that LLMs can be utilized for automatic evaluation of attribution, operationalizing the guidelines presented in \cite{rashkin2023measuring}. However, both of these works are primarily designed for the question-answering (QA) task. In contrast, our method is not restricted to QA and is designed to work with text in general. Furthermore, while these previous studies focus on sentence or paragraph levels, our approach extends to a more detailed and granular level of analysis.

\begin{figure*}[h]
    \centering
    \includegraphics[width=0.95\linewidth]{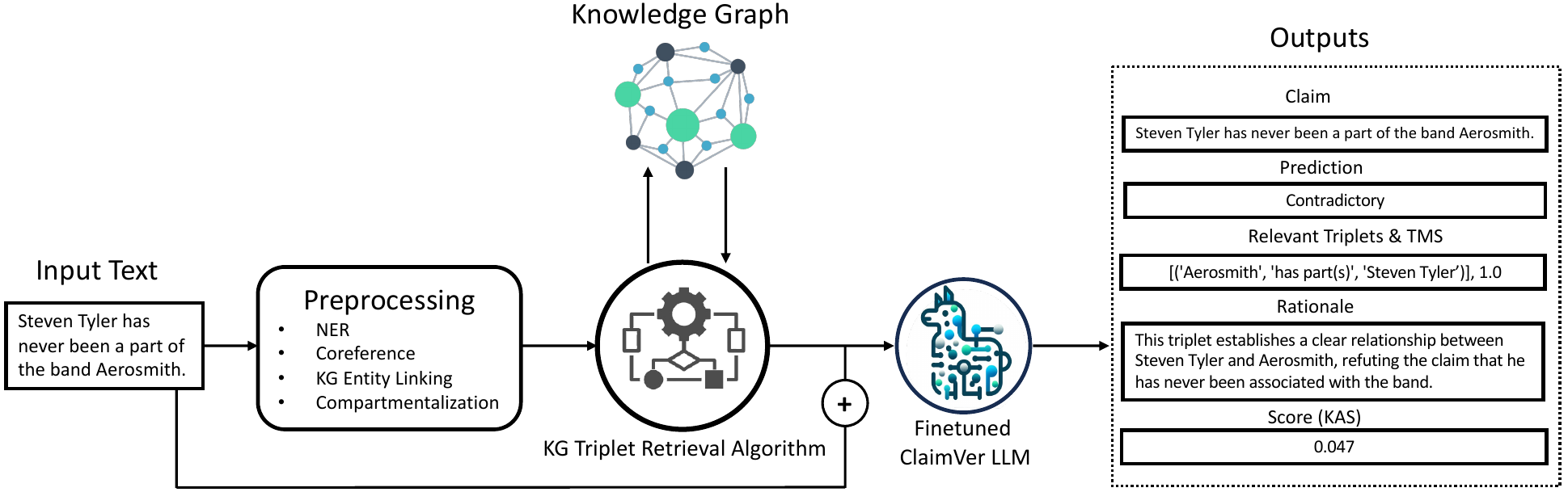}
    \caption{Flow of operations in the ClaimVer framework. Identified KG entity nodes during preprocessing inform the extraction of relevant triplets by the KG algorithm. Subsequently, these triplets and preprocessed text are then fed to a ClaimVer LLM, fine-tuned to operationalize the objective function. For each claim, the corresponding text span, prediction, relevant triplets, attribution scores, and rationale are generated.}
    \label{fig:pipeline}
    \vspace{-5mm}
\end{figure*}

\section{Methodology}
\label{sec:method}
In this section, we present the methodology for retrieving relevant triplets from the KG, fine-tuning LLM to process text at claim-level, verifying claims, tagging evidence for each prediction, and generating a rationale along with an attribution score that reflects the text's validity.
%We discuss the approach in logical parts and tie everything in \S\ref{sec:flowOfOp}.

\subsection{Preprocessing}
\label{sec:preprocessing}
Preprocessing involves multiple steps required to make the input text suitable for the subsequent operations. Since the nodes in a KG typically represent entities, performing Named Entity Recognition (NER) is necessary. In our work, we chose Wikidata \cite{vrandevcic2014wikidata} as the KG source; thus, we use an NER module suitable for Wiki entities \cite{spaCyEntityLinker}. However, the framework is sufficiently generic to support any kind of KG that models information in the form of triplets. As our analysis is performed at the claim level, coreference resolution \cite{lee2017end} becomes a necessary step to form localized claims that are semantically self-contained. If input text exceeds the context length, which depends on design choices, compartmentalization would be required. As a final step in preprocessing, we perform KG entity linking.  This step tags all entities in the text that are present in the KG as nodes.

\subsection{Relevant Triplets Retrieval}
Retrieving relevant triplets is a complex problem that has attracted attention from various research communities, and resulted in multiple approaches to address the challenge. While retrieving direct links between two given nodes in a KG is relatively straightforward, identifying complex paths that involve multiple hops is challenging. In our framework, we use Woolnet \cite{Gutierrez2023-ts}, a multi-node Breadth-First Search (BFS) algorithm, to retrieve the most relevant triplets for a given claim present in the KG.
This BFS algorithm initiates from multiple starting points and, at each step, searches for and processes all adjacent neighbors before advancing. It constructs a subgraph of visited nodes, tracking their origins, and distances from each BFS's start. The algorithm expands each search tree one node at a time until paths intersect or reach a predefined maximum length. Upon intersection, it assesses if the discovered path meets the length criteria. If so, it logs the route, utilizing backtracking to trace the path to its origins, while ensuring there are no repetitions or cycles, thus maintaining a connection to a starting node. In our experiments, we allow for a maximum of three hops between any two given nodes, and a maximum of four potential paths. Adopting less stringent conditions leads to less relevant triplets.

\subsection{Objective Function}
\label{sec:AttrObjFunc}
Previous works on evidence attribution tasks have established definitions for the categorization of input text with reference to a supporting source \cite{rashkin2023measuring,gao2023rarr,bohnet2022attributed,yue2023automatic}. Similar to the formulation in \cite{yue2023automatic}, we use three categories: \textit{Attributable}, \textit{Extrapolatory}, and \textit{Contradictory}. However, there are two main differences that distinguish our approach from previous methods. First, we verify the input text against facts present in a KG, an aggregated information source constructed by integrating numerous data sources into a structure of triplets, instead of relying on a single reference. This approach eliminates the one-to-one dependency between the text and its information source. Second, we perform attribution with finer granularity, specifically at the claim level, involving a subtask of decomposing the input text into individual claims. We define our categories as follows:

\begin{itemize}
\item \textbf{Attributable}: Triplets fully support the claim.
\item \textbf{Extrapolatory}: Triplets lack sufficient information to evaluate the claim.
\item \textbf{Contradictory}: Triplets contradict the claim.
\end{itemize}

We formulate the objective function of our task as follows:
\begin{equation}
\scalebox{0.9}{$
\begin{aligned}
    f(&input\_text, ret\_triplets) = \\
    &\{(claim\_span_i, claim\_pred_i, \\
    &rel\_triplets_i, rationale_i) \}_{i=1}^{n}
\end{aligned}
$}
\end{equation}

where:
\begin{itemize}[left=0pt]
    \item \parbox[t]{1\linewidth}{$input\_text$: input text containing claim(s).}
    \item \parbox[t]{1\linewidth}{$ret\_triplets$: \textit{retrieved} triplets for the input text.}
    \item \parbox[t]{1\linewidth}{$claim\_span_i$: $i^{th}$ claim extracted as a substring from $input\_text$.}
    \item \parbox[t]{1\linewidth}{$claim\_pred_i$: label predicted for $claim\_span_i$.}
    \item \parbox[t]{1\linewidth}{$rel\_triplets_i$: \textit{relevant} subset of $ret\_triplets$ for $claim\_span_i$.}
    \item \parbox[t]{1\linewidth}{$rationale_i$: justification for $claim\_pred_i$.}
    \item \parbox[t]{1\linewidth}{$n$: total number of claims in $input\_text$.}
\end{itemize}

This objective function encompasses two main sub-tasks:
\begin{enumerate}
    \item Decomposing input text into claims.
    \item Generating prediction and corresponding rationale for each claim by identifying relevant supporting triplets.
\end{enumerate}

\subsection{Fine-tuning LLMs}
\label{sec:fine-tuning}

The objective function shares similarities with the well-studied task of NLI \cite{conneau2017supervised,schick2020exploiting}. LLMs achieve state-of-the-art performance for NLI \cite{chowdhery2023palm}, making them a suitable choice to operationalize the objective function. Additionally, \cite{yue2023automatic} shows that LLMs can be used to automatically evaluate attribution to a given information source. However, these prior methods do not involve a complex sub-task, which is central to the proposed objective function, i.e., decomposing the input text into text spans that correspond to separate claims in the presence of multiple claims.

It is crucial to perform both claim decomposition and attribution for all claims in a single step, as processing each claim individually can lead to an exponential increase in LLM queries, leading to significantly higher computational costs and latency issues. 
% To achieve this new task, we finetune LLMs specifically for the objective function described in \S\ref{sec:AttrObjFunc}.

In order to perform attribution at the claim level, we need to fine-tune LLMs specifically for the proposed objective function (see \S\ref{sec:AttrObjFunc}) using a custom dataset. This is necessary because, as of this writing, even the state-of-the-art model, OpenAI's GPT-4 \cite{achiam2023gpt}, does not perform satisfactorily right out of the box. Our custom dataset, built using two sequential complex prompts with GPT-4, enables us to fine-tune significantly smaller models. This approach distills the performance of a large proprietary model using a multi-query prompt pipeline into small open-source models with a compact zero-shot prompt. We make the weights of the fine-tuned models publicly available\footnote{\href{https://huggingface.co/preetam7}{weights available on HuggingFace}}.

We selected eight open-source LLMs with diverse sizes, ranging from 2B parameters to 10B parameters, to perform the fine-tuning: Gemma-2B-IT-Chat \cite{team2024gemma}, Phi-3-mini-4k-Chat \cite{javaheripi2023phi}, Zephyr-7B-Beta-Chat \cite{tunstall2023zephyr}, Mistral-7B-v0.3-Chat \cite{jiang2023mistral},  Llama3-8B-Chat \cite{touvron2023llama}, and Solar-10.7B-Chat \cite{kim2023solar}. The models were fine-tuned using LoRA \cite{hu2021lora} with 4-bit quantization and adapters with rank 8 \cite{dettmers2024qlora}. The context length was set to 4096 tokens (for additional training details, refer \S\ref{app:training}) All models converged after 2 epochs, and high ROUGE-L \cite{lin-2004-rouge} scores greater than 0.658 were achieved for each model. The instruction prompt used for fine-tuning is presented in Figure \ref{fig:analysis_prompt}.

\setlength{\fboxsep}{3pt}
\setlength{\fboxrule}{1pt}

\begin{figure}[t] % Use [ht] for "here" or "top" placement of the float
\centering % Centers the minipage on the page
\fbox{%
\begin{minipage}{0.99\columnwidth} % Makes the minipage the width of the column
\scriptsize % Sets a smaller font size
\texttt{%
Analyze text against provided triplets, classifying claims as "Attributable", "Contradictory", or "Extrapolatory".\newline
Justify your classification using the following structure:\newline
- "text\_span": Text under evaluation.\newline
- "prediction": Category of the text (Attributable / Contradictory / Extrapolatory).\newline
- "triplets": Relevant triplets (if any, else "NA").\newline
- "rationale": Reason for classification.\newline
For multiple claims, number each component (e.g., "text\_span1", "prediction1",..). Use "NA" for inapplicable keys.\newline
Example:\newline
"text\_span1": "Specific claim",\newline
"prediction1": "Attributable/Contradictory/Extrapolatory",\newline
"triplets1": "Relevant triplets",\newline
"rationale1": "Prediction justification",\newline
...\newline
Input for analysis:\newline
-Text: \{Input Text\}\newline
-Triplets: \{Retrieved Triplets\}\newline
}
\vspace{-2mm}
\end{minipage}
}
\caption{Instruction prompt for fine-tuned LLMs.}
\label{fig:analysis_prompt}
\vspace{-5mm}
\end{figure}

\subsection{Computing Attribution Scores}

For various downstream tasks, such as ranking and filtering, a continuous score that reflects the validity of a given piece of text with respect to a KG is desirable. We propose the KG Attribution Score (KAS), which accomplishes this task with a high level of granularity, and is detailed in this section.

\subsubsection{Claim Scores}
\begin{equation}
\scalebox{0.8}{$
\text{cs}(y_i) = 
\begin{cases} 
2 & \text{if } y_i = \text{Attributable} \\
1 & \text{if } y_i = \text{Extrapolatory} \text{ and } |\textit{rel\_triplets}_i| > 0 \\
0 & \text{if } y_i = \text{Extrapolatory} \text{ and } |\textit{rel\_triplets}_i| = 0 \\
0 & \text{if } y_i = \text{No attribution} \\
-1 & \text{if } y_i = \text{Contradictory}
\end{cases}
$}
\end{equation}
\text{where, } $y_i$ \text{ is } $claim\_pred_i$.

For each claim, we assign a score that reflects the level of its validity, ranging from -1 (\textit{contradictory}) to 2 (\textit{attributable}). If a claim is predicted to be \textit{extrapolatory}, yet has one or more relevant triplets, we assign that claim a score of 1, as there is still relevant information available even though it may not be sufficient to completely support or refute the claim. However, if there are no triplets at all, along with an \textit{extrapolatory} prediction, we assign 0 as it does not add any useful information. While decomposing claims, the model might occasionally omit words, typically stop-words, and we assign 0 in those cases as well.

\subsubsection{Triplets Match Score (TMS)}
This score reflects the extent of the match between the relevant triplets and the corresponding claim, and it can also serve as a proxy for the prediction confidence. Even though the prediction is made at the claim level, the triplets match score considers word-level matches in the computation. It can be computed as follows:

\begin{equation}
\scalebox{0.79}{$
\begin{aligned}
    TMS(&E(claim\_span_i), E(rel\_triplet_i)) = \\
    &\alpha \cdot SS(E(claim\_span_i), E(rel\_triplet_i)) \\
    &+ \beta \cdot EPR(E(claim\_span_i), E(rel\_triplet_i))
\end{aligned}
$}
\end{equation}

where, $E(claim\_span_i)$ and $E(rel\_triplet_i)$ represent the sets of entities in $claim\_span_i$ and $rel\_triplet_i$, respectively. $SS$ is the semantic similarity computed using the cosine similarity of text embeddings, and $EPR$ represents the ratio of entities in $E(claim\_span_i)$ that are also present in $E(rel\_triplet_i)$. The parameters $\alpha$ and $\beta$ can be adjusted as needed; in our experiments, we use 0.5 for both. In cases where examples of an entity retrieved from the KG are used to support the prediction, instead of the entity itself, we may not have a direct overlap, and thus semantic similarity would be helpful. $EPR$ rewards the direct use of the entity, so a balance between both may be ideal in most cases.

\subsubsection{KG Attribution Score (KAS)}

For the final KG Attribution Score (KAS), a continuous score between 0 and 1 is desirable, as this facilitates various downstream applications such as ranking, fine-tuning, and filtering. This can be achieved using a Sigmoid function. However, the standard Sigmoid function treats positive and negative scores equally. In most cases, higher penalties should be assigned for erroneous text than rewards for valid text. This requirement can be met using a modified Sigmoid function that penalizes mistakes by a factor of $\gamma$:

\begin{equation}
\begin{aligned}
\sigma_{\text{mod}}(x, \gamma) = \frac{1}{1 + e^{-\gamma \cdot x}}, \\
\text{where } \gamma = 
\begin{cases} 
\gamma=3 & \text{if } x < 0, \\
\gamma=1 & \text{if } x \geq 0,
\end{cases}
\end{aligned}
\end{equation}

In our experiments, we set the value of $\gamma$ to 3. Finally, the modified Sigmoid function, applied to the summation of triplet match scores and claim scores, is used to generate KAS:

\begin{equation}
\begin{aligned}
    \text{KAS} = \sigma_{mod} (\sum_{i=1}^{n} [&TMS_i \cdot cs(y_i)], \gamma)
\end{aligned}
\end{equation}

% \subsection{Flow of Operation}
% \label{sec:flowOfOp}

\section{Dataset}
\label{sec:dataset}

\begin{table}[t]

\begin{center}
\centering
\resizebox{0.75\columnwidth}{!}{   
\begin{tabular}{@{}lccccc@{}}
\toprule
\multicolumn{1}{c}{\multirow{2}{*}{\textbf{Split}}} & \multirow{2}{*}{\textbf{Samples}} & \multirow{2}{*}{\textbf{Claims}} & \multicolumn{3}{c}{\textbf{Claim Labels}}  \\ \cmidrule(l){4-6} 
\multicolumn{1}{c}{}                                &                                   &                                  & \textbf{Att} & \textbf{Ext} & \textbf{Con} \\ \midrule
Train                                               & 3400                              & 5342                             & 998         & 3546         & 798          \\
Test                                                & 1000                              & 1677                             & 316          & 1068          & 293          \\ \bottomrule
\end{tabular}
}
\caption{Distribution of fine-tuning dataset. Att: Attributable, Ext: Extrapolatory, Con: Contradictory.  \label{table:dataset}}
\end{center}
\vspace{-5mm}
\end{table}
% Counter({'Extrapolatory': 1068, 'Attributable': 316, 'Contradictory': 293}) test
% 3546+998+798

\begin{table*}[h]
    \centering
    \includegraphics[width=1\linewidth]{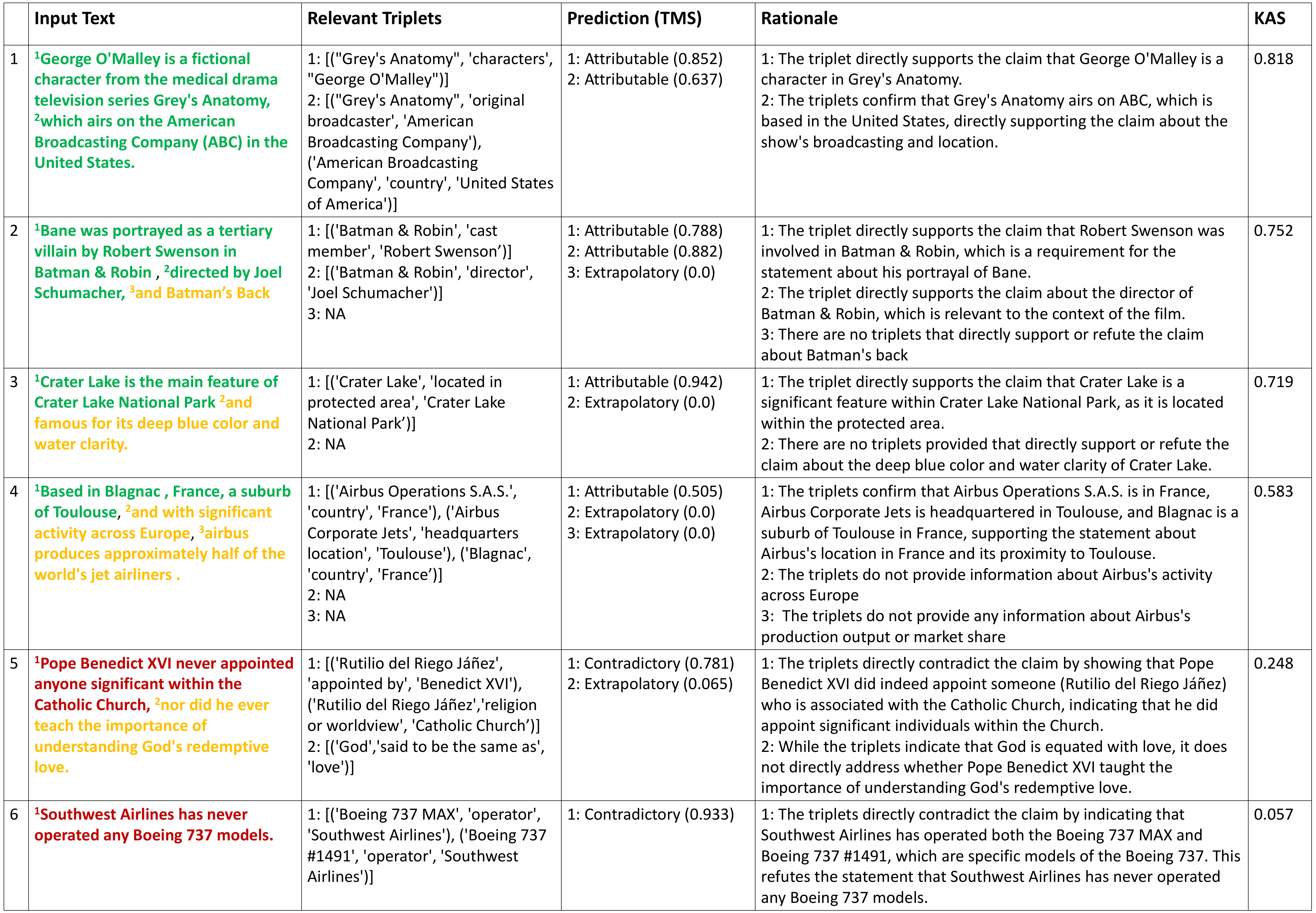}
    \caption{Examples of claim-level attribution by the proposed method. The first column shows the numbered claims in the input text. Second column lists relevant triplets for each claim. Predictions and \textit{Triplets Match Score (TMS)} are in the third column, while the rationale behind each prediction is in the fourth column. The \textit{Knowledge Graph Attribution Score (KAS)} is shown in the last column. Model: \textit{Solar-10.7B-Chat}. \label{fig:examplesTable}}
    \vspace{-5mm}
\end{table*}

% Satisfactory performance was achieved through extensive prompt tuning and two large sequential prompts were required. Further details on the dataset are provided in \S\ref{sec:dataset}. 

Open-domain Question Answering (QA) datasets, such as WikiQA \cite{yang-etal-2015-wikiqa}, HotPotQA \cite{yang2018hotpotqa}, PopQA \cite{mallen2022not}, and EntityQuestions \cite{sciavolino-etal-2021-simple}, as well as Fact Verification datasets like FEVER \cite{thorne-etal-2019-fever2}, FEVEROUS \cite{aly2021feverous}, TabFacT \cite{chen2019tabfact}, and SEM-TAB-FACTS \cite{wang2021semeval}, provide texts along with corresponding reference contexts or attributable information sources. However, these datasets significantly differ from the type of data required to train and test our proposed objective function, primarily due to two major factors: (i) these datasets predominantly offer samples that are inherently \emph{attributable}, and (ii) consist of atomic claims and/or one-to-one mappings between input and reference texts. To address the first limitation, prior work \cite{yue2023automatic} in attribution evaluation introduced new samples by modifying correct answers to generate \emph{contradictory} instances. Yet, this adjustment alone is not sufficient for our use case because our method requires attribution at the claim level, and necessitates the automatic decomposition input text to claims. Consequently, as this task represents a novel challenge, we developed a new dataset that enables effective training and testing of the objective function. 

Considering the choice of our KG, which is Wikidata \cite{vrandevcic2014wikidata}, we opted for WikiQA \cite{yang-etal-2015-wikiqa} as it is closely associated with the Wiki ecosystem. Given that our method is designed for text validation in general, and is not limited to question answering, we retain only answers and discard the questions. Subsequently, we processed the answers following the steps detailed in Section \ref{sec:preprocessing}, selecting entries containing two or more Wiki entities. This approach resulted in the exclusion of most single-word answers and other responses that are dependent on their corresponding questions and may lack comprehensibility without them.

We utilize GPT-4 \cite{achiam2023gpt} to generate the initial version of the ground truth. Although GPT-4 can adhere to the instructions (refer to Figure \ref{fig:analysis_prompt}) to a reasonable degree and responds in the required format with all necessary keys, it still underperforms in the overall task. The most frequent issue observed is the erroneous assignment of prediction labels. To remedy this issue, we designed a detailed prompt tailored for the given task, incorporating techniques such as few-shot, chain-of-thought \cite{kojima2022large}, and other strategies \cite{openai_promptengineering,nori2023can} (full prompt in \S\ref{appendix} Figure \ref{fig:App_prompt2}).
We also conducted manual checks to ensure only high-quality samples were retained, as research indicates that high alignment can be achieved with as few as 1,000 samples, provided they are of superior quality \cite{zhou2023lima}.

The final dataset is comprised of two splits: the training split, based on the training split of WikiQA \cite{yang-etal-2015-wikiqa}, and a test split, derived from both the test and validation splits. The training split contains 3,400 samples, and since some entries feature multiple claims, there are a total of 5,342 claims within this split. Similarly, the test split includes 1,000 samples and 1,677 claims. The label counts for the claims are tabulated in Table \ref{table:dataset}. The dataset is publicly shared to facilitate further research in this direction\footnote{\href{https://huggingface.co/datasets/preetam7/ClaimVer}{dataset available on HuggingFace}}.

\section{Experiments and Results}

In this section, we present the evaluation of our claim-level attribution method. The performance metrics of the fine-tuned LLMs, which operationalize the objective function, are presented in Tables \ref{table:rouge} and \ref{table:claimScores}. In Table \ref{table:rouge}, we observe that all models converge and achieve sufficiently high ROUGE-L and ROUGE-1 scores, with \textit{Mistral-7B-v0.3-Chat} achieving the highest of 0.694 and 0.719 respectively. We also observe that the smaller model, \textit{Gemma-2B-IT-Chat} with just 2B parameters, is also sufficiently compatible for this task as it attained a decent ROUGE-L score of 0.667.

% The reason behind this variation is that these scores account for both sub-tasks of the objective function (refer \S\ref{sec:AttrObjFunc}): decomposing input text into claims and generating predictions for each claim. 

The first task of the proposed objective function  (refer \S\ref{sec:AttrObjFunc}), decomposing text into multiple claims, is somewhat subjective, and there could be multiple valid approaches due to linguistic complexities. For instance, example 4 in Table \ref{fig:examplesTable} has been decomposed into three claims, but the first could arguably be further decomposed to verify whether Blagnac is in France, and whether it is a suburb of Toulouse. Controlling the precise manner of decomposition is challenging, and might necessitate an additional step before the prediction step, involving separate processing for each claim. However, this option could prove to be impractical, as the number of LLM queries could increase exponentially. 

% To verify focusing solely on the prediction task could offer better insights into how the models are performing in terms of categorization.

To accurately compute classification performance, we impose a strict strategy: the text span of the claim, the identified relevant triplets, and the prediction label must all exactly match the ground truth to be considered accurate. In Table \ref{table:claimScores}, the second column indicates number of claims with text spans exactly matching the ground truth responses. Columns 3 to 6 present the accuracy, precision, recall, and F1 scores for these matching claims. The most performant model is \textit{Solar-10.7B-Chat}, with 1031 exact matches out of 1677 claims in the test set. Additionally, the classification scores in all metrics are above 89\%, which clearly demonstrates that the model can reliably differentiate between the classes \textit{attributable}, \textit{extrapolatory}, and \textit{contradictory}.

Table \ref{fig:examplesTable} showcases the claim-level attribution performed by our method. Each claim in the input text is numbered and color-coded to reflect its prediction: green for attributable, amber for extrapolatory, and red for contradictory. The examples are sorted in descending order by their KAS scores, which reflect the validity of the text. As expected, we observe more green at the top of the table and more amber and eventually red as we move down. Since the Wiki ecosystem is open-domain, we observe that the examples cover a wide range of topics, demonstrating that the method is adaptable to diverse inputs.

In the first example in \ref{fig:examplesTable}, the input text is decomposed into two claims, both of which are attributable. The first claim is supported by a single triplet in the KG, while the second claim can be supported by combining two triplets. The second example presents more challenges for evaluation due to its complex sentence structure, but ClaimVer accurately identifies that the third claim regarding Batman's Back is neither supported nor refuted by the triplets, as indicated in the rationale. In the third example, we note that the first claim is predicted to be attributable with a high triplet match score of 0.942 since there is a triplet that clearly supports the location description of Crater Lake. However, as there is no information regarding the water characteristics, the second claim is categorized as extrapolatory. In the fourth example, the first claim alone requires three triplets combined as supporting evidence, illustrating the method's ability to handle complex multi-hop paths within the KG. The second and third claims are predicted to be extrapolatory, since there are no triplets concerning Airbus's market share, or its activities in Europe, as highlighted in the model's rationale. It is noteworthy that the context provided in the third claim is crucial for the first claim to be comprehensible, demonstrating why individual claim evaluation may be suboptimal. Interestingly, in the fifth example, the method identifies a specific instance from the KG to refute a general claim, citing the appointment of Rutilio del Riego Jáñez. Similarly, in the sixth example, the method provides specific instances, quoting two distinct Boeing 737 models to demonstrate contradiction with a high triplet match score.

% \begin{table}[t]
% \centering
% \resizebox{0.85\columnwidth}{!}{   
% \begin{tabular}{@{}lcccc@{}}
% \toprule
% \textbf{Model}   & \textbf{Size} & \multicolumn{1}{l}{\textbf{ROUGE-L}} & \multicolumn{1}{l}{\textbf{ROUGE-1}} & \multicolumn{1}{l}{\textbf{Avg. Acc.}} \\ \midrule
% Phi-2            & 2.7B          & 0.635                                & 0.673                                & 75.16\%                               \\
% Mistral-Instruct & 7B            & 0.645                                & 0.680                                & 83.04\%                               \\
% Zephyr-Beta      & 7B            & 0.638                                & 0.676                                & 79.88\%                               \\
% Solar-10.7B-Chat   & 10.7B         & \textbf{0.655}                       & \textbf{0.693}                       & \textbf{89.31\%}                      \\
% Llama2-Chat      & 13B           & 0.6395                               & 0.677                                & 79.41\%                               \\ \bottomrule
% \end{tabular}
% }
% \caption{ROUGE scores and average accuracies on the test set ($n=1,000$). \label{table:rouge}}
% \end{table}

\begin{table}[t]
\centering
\resizebox{0.8\columnwidth}{!}{   
\begin{tabular}{@{}lccc@{}}
\toprule
\textbf{Model}                       & \textbf{Size} & \multicolumn{1}{l}{\textbf{ROUGE-L}} & \multicolumn{1}{l}{\textbf{ROUGE-1}} \\ \midrule
Gemma-2B-IT-Chat    & 2B            & 0.667                                & 0.692                                \\
Phi-3-mini-4k-Chat      & 4B            & 0.658                                & 0.685                                \\
Zephyr-7B-Beta-Chat       & 7B            & 0.686                                & 0.712                                \\
Vicuna-7B-v1.5-Chat    & 7B            & 0.676                                & 0.700                                \\

Mistral-7B-v0.3-Chat   & 7B            & \textbf{0.694}                                & \textbf{0.719}                                \\
Gemma-7B-IT-Chat    & 7B            & 0.678                                & 0.703                                \\
Llama3-8B-Chat      & 8B            & 0.679                                & 0.705                                \\
Solar-10.7B-Chat       & 10B           & 0.689                                & 0.714                                \\ \bottomrule
\end{tabular}
}
\caption{ROUGE scores on the test set ($n=1,000$). \label{table:rouge}}
\end{table}

\begin{table}[t]
\centering
\resizebox{0.8\columnwidth}{!}{   
\begin{tabular}{@{}lccccc@{}}
\toprule
\textbf{Model}                     & \textbf{\#MC} & \textbf{Acc}  & \textbf{Prec} & \textbf{Rec}  & \textbf{F1}    \\ \midrule
Gemma-2B-IT-Chat                   & 895           & 77.09         & 77.20         & 77.09         & 74.24          \\
Phi-3-mini-4k-Chat                 & 882           & 72.22         & 78.10         & 72.22         & 72.86          \\
Zephyr-7B-Beta-Chat                & 978           & 85.89         & 87.41         & 85.89         & 86.16          \\
Vicuna-7B-v1.5-Chat                & 898           & 79.62         & 78.83         & 79.62         & 78.84          \\
Mistral-7B-v0.3-Chat               & 1002          & 86.63         & 87.03         & 86.63         & 86.73          \\
Gemma-7B-IT-Chat                   & 940           & 82.87         & 84.09         & 82.87         & 83.17          \\
Llama3-8B-Chat                     & 959           & 80.92         & 85.48         & 80.92         & 81.36          \\
Solar-10.7B-Chat                   & \textbf{1031}          & \textbf{89.23}         & \textbf{89.52}         & \textbf{89.23}         & \textbf{89.30}          \\ \bottomrule
\end{tabular}
}
\caption{Scores on matching claims in the test set ($n=1677$). \#MC: number of matching claims. 
\label{table:claimScores}}
\end{table}

% \begin{table}[t]
% \centering
% \resizebox{0.8\columnwidth}{!}{   
% \begin{tabular}{@{}lccccc@{}}
% \toprule
% \textbf{Model}   & \textbf{\#MC} & \textbf{Acc}   & \textbf{Prec}  & \textbf{Rec}   & \textbf{F1}    \\ \midrule
% Phi-2            & 819           & 98.29          & 98.33          & 98.29          & 98.26          \\
% Mistral-Instruct & 928           & 99.35          & 99.36          & 99.35          & 99.35          \\
% Zephyr-Beta      & 757           & 98.34          & 98.38          & 98.34          & 98.32          \\
% Solar-10.7B-Chat   & \textbf{1052} & \textbf{99.80} & \textbf{99.81} & \textbf{99.80} & \textbf{99.80} \\
% Llama2-Chat      & 869           & 99.53          & 99.54          & 99.53          & 99.53          \\ \bottomrule
% \end{tabular}
% }
% \caption{Scores on matching claims in the test set ($n=1677$). \#MC: number of matching claims. 
% \label{table:claimScores}}
% \end{table}

\section{Discussion}
\label{sec:discussion}

The susceptibility of LLMs to generating factually incorrect statements is an alarming concern as LLM-powered services become increasingly popular for seeking advice and information. The democratization of generative models has also had adverse effects, such as increasing misinformation \cite{monteith2024artificial}. To arm end-users with the tools necessary to combat being misinformed, it is crucial to develop text-validation methods that are human-centric, and prioritize user engagement, understanding, and informativeness. We design our method with these principles in mind: we make predictions at the claim level, and identify text spans within the given text, that can be color-coded and presented to the user. The proposed method also generates easily comprehensible explanations along with the prediction and evidence, thus reducing the cognitive burden on the end-user, and making the process user-friendly.

The usability and evaluation of these systems should align with human needs and capabilities. Chatbots, such as ChatGPT \cite{achiam2023gpt}, serve a wide array of tasks; therefore, the text validation method should be adaptable to various domains. While KGs like Wikidata \cite{vrandevcic2014wikidata} are considered open-domain, the implementation of more specialized KGs, along with corresponding routing algorithms may be necessary to support a broader range of topics. For instance, a common-sense KG \cite{Hwang2020COMETATOMIC2O} would be more useful in validating non-factoid answers that involve logic. 

Furthermore, the maintenance efficiency of our approach aligns well with the need for sustainable, long-term AI solutions. In a world where information is constantly evolving, the ability to update and maintain AI systems with minimal effort is not just a convenience, but a necessity. This directly ties into the ethical implications of AI, where outdated or incorrect information can lead to harmful decisions. By leveraging existing, well-maintained KGs, we can ensure that AI systems remain accurate and relevant over time.

\section{Conclusion}

In this paper, we present ClaimVer, a framework for text verification and evidence attribution at the claim level by leveraging information present in KGs. In contrast to other methods, ClaimVer eliminates the one-to-one mapping between input and reference text, allowing for layered interpretation and handling of distributed information. In addition to these primary functions, ClaimVer incorporates human-centric design principles by offering clear, concise explanations for each claim prediction—an important characteristic for building user trust and enhancing usability. Furthermore, we introduce an attribution score, which enhances its applicability across a wide range of downstream tasks. Finally, we share ClaimVer fine-tuned LLMs to facilitate further exploration of this research direction.

% We have prioritized human-centric design principles to make the framework more informative, intuitive, and user-friendly. Additionally, our methodology incorporates design choices that ensure open access, sustainability, and reliability.

% ClaimVer presents several advantages, as outlined below:
% \begin{enumerate}
% \item \textit{Human-centric design:}  In addition to its primary functions of text verification and evidence attribution, the system generates considerable information conducive to user awareness. This information serves to educate users and enhance their trust in the automated system.
% \item \textit{Finer Granularity:} Perform validation at the claim level, enabling localization of hallucinations, or false claims.
% \item \textit{Enhanced Coverage:} Eliminate one-to-one mapping between input and reference text, allowing for layered interpretation, and handling of distributed information.
% \item \textit{Domain Adaptability:} Flexibility in adapting to new domains by switching to a more suitable KG.
% \item \textit{Maintenance Efficiency:} Simplified auditing and updation of the knowledge base, ensuring the data remains current and accurate.
% \end{enumerate}

\section{Limitations}

\noindent{\textbf{Limitations of LLMs for Fact Verification.}} Like most ML models, LLMs are prone to erroneous predictions, which is particularly concerning in sensitive applications such as handling misinformation. Despite this, they remain the most performant techniques for fact verification and related tasks like NLI \cite{yue2023automatic,wang2021entailment}. Therefore, while it is reasonable to use the best option available, fact verification systems relying on LLMs should be utilized with caution and necessary validations.

\noindent{\textbf{Limitations of Knowledge Graphs.}}
While there are several advantages associated with using KGs, we also acknowledge the presence of known issues, such as knowledge coverage and the efforts required to keep these sources up-to-date. For our solution, we assume that the KG is up-to-date and possesses adequate coverage. However, this may not always be the case, and thus the most suitable technique should be adopted after considering the specific requirements of a particular use case. Another point to consider is that the proposed method does not provide traditional citations to articles, although it may be possible to retrieve that information from the KG, if information source mapping has been properly maintained.

\noindent{\textbf{Variations in Claim Decomposition.}} Decomposing text into multiple claims is a complex linguistic task that often results in multiple valid decompositions. Although this may not impact usability if the prediction, rationale, and text spans are comprehensible and supported by facts from the KG, it poses a challenge for evaluating model performance. One potential approach is to operate at the token level instead of the span level, but this would significantly complicate the problem space. Additionally, token-level verification and attribution would require substantially higher compute resources, necessitating further studies to assess their value and impact on system usability and reliability.

\noindent{\textbf{LLM Reasoning Errors.}} Previous works have demonstrated that using LLM reasoning for tasks like fact verification, evidence attribution, and NLI can yield impressive results, surpassing alternative approaches \cite{yue2023automatic,wang2021entailment}. However, LLM reasoning can still be flawed. In our work, we impose validations to minimize LLM mistakes by performing membership checks for supporting triplets and string matching for text spans. Yet, validating reasoning remains an open problem with ongoing research efforts.

\noindent{\textbf{Fine-tuning Dataset Limitations.}} To build the fine-tuning dataset, we utilized GPT-4 with two detailed sequential prompts designed in accordance with OpenAI's recommendations \cite{openai_promptengineering} and previous works \cite{nori2023can}. Despite employing techniques like few-shot prompting with state-of-the-art LLMs, we still observe mistakes, indicating the complexity of this problem. To address this, we conducted manual checks to minimize errors and share the dataset with the research community for further improvement.

\section{Ethical Concerns}
Our work presents a scalable and interpretable framework for fact-checking textual claims. To promote the exploration of this important problem space, we fine-tune and share open-source LLMs that are well-aligned to the framework's objective function. While the models we provide perform better than the publicly available base models for our specific task, they still share similar weaknesses such as erroneous reasoning. To address this as best as we can, we have incorporated and described ways to mitigate these issues to the extent possible. We believe the benefits of this work outweigh any potential risks.

% We utilize fine-tuning to distill the performance of a large proprietary LLM using two detailed sequential prompts into smaller open-source models with more compact zero-shot prompts. Despite employing techniques like few-shot prompting with state-of-the-art LLMs, we still observe mistakes, indicating this is not a solved problem. To address this, we conducted manual checks to minimize errors and share the dataset with the research community for further improvement.

% Bibliography entries for the entire Anthology, followed by custom entries
%\bibliography{anthology,custom}
% Custom bibliography entries only
\bibliography{acl_latex}

\newpage
\appendix
\section{Appendix}
\label{appendix}

\subsection{Training Details}
\label{app:training}
In this section, we present the training parameters used for fine-tuning each model, along with their corresponding loss plots. All models converged after two epochs, achieving ROUGE-L \cite{lin-2004-rouge} scores greater than 0.658, with the best model reaching 0.719.

\begin{figure}[h]
\centering
  \includegraphics[width=1\linewidth]{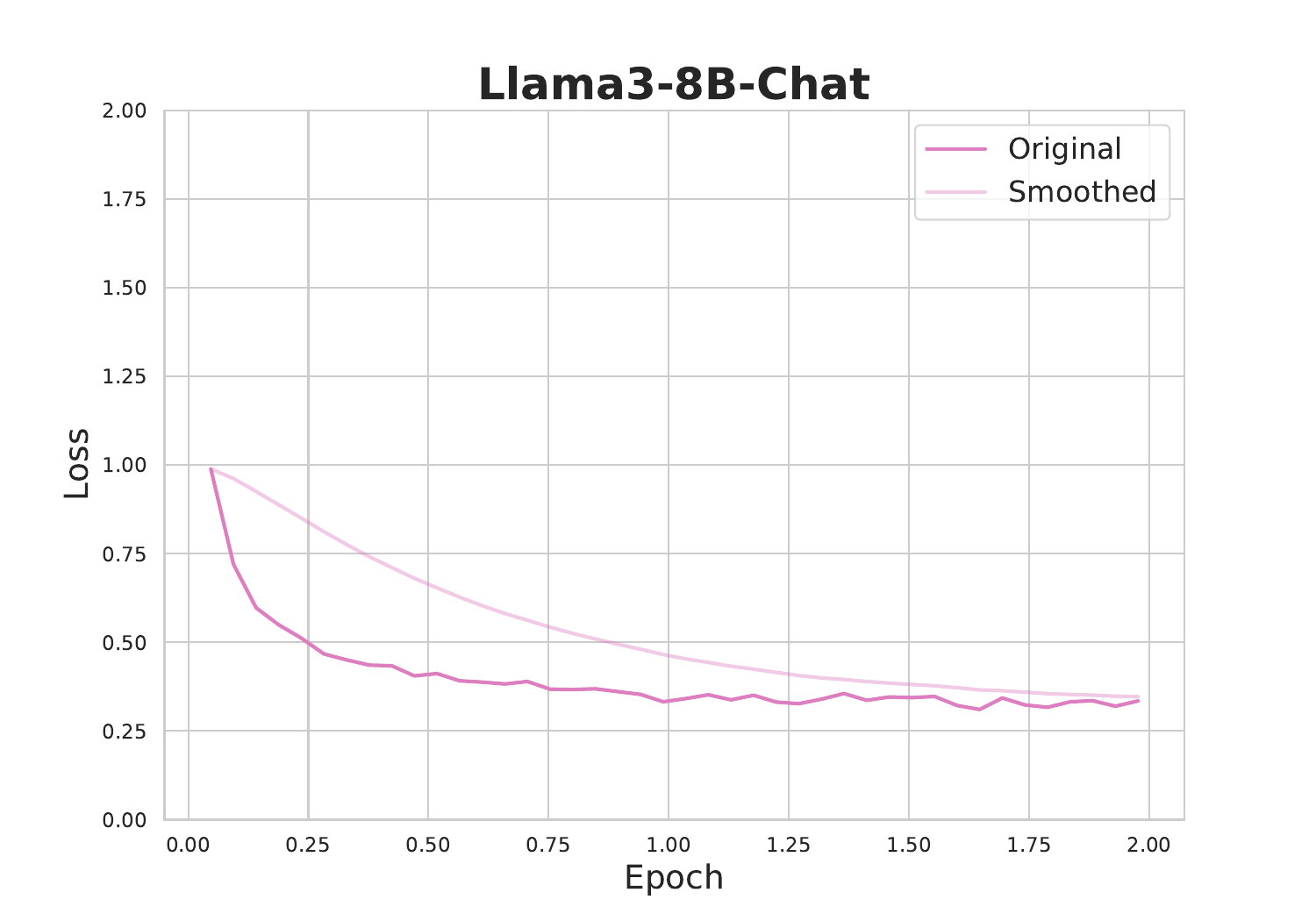}
  \caption{\small{Fine-tuning loss plots for Llama3-8B-Chat.}}
\end{figure}

\begin{table}[h]
    \centering
    \resizebox{1\columnwidth}{!}{   
    \begin{tabular}{ll}
        \toprule
        \textbf{Parameter} & \textbf{Value} \\ \midrule
        Base Model & meta-llama/Meta-Llama-3-8B-Instruct \\ 
        ROUGE-L & 0.679 \\ 
        ROUGE-1 & 0.705 \\ 
        Fine-Tuning Type & LoRA \\ 
        LoRA Alpha & 16 \\ 
        LoRA Rank & 8 \\ 
        Cutoff Length & 4096 \\ 

        Gradient Accumulation Steps & 8 \\ 
        Learning Rate & 5.0e-05 \\ 
        LR Scheduler Type & Cosine \\ 
        Number of Training Epochs & 2.0 \\ 
        Optimizer & AdamW \\ 
        
        Quantization Bit & 4 \\ 
        \bottomrule
    \end{tabular}
    }
    \caption{Fine-tuning Parameters for Llama3-8B-Chat}
    \label{tab:training_params_Llama3-8B-Chat}
\end{table}

\begin{figure}[h]
\centering
  \includegraphics[width=1\linewidth]{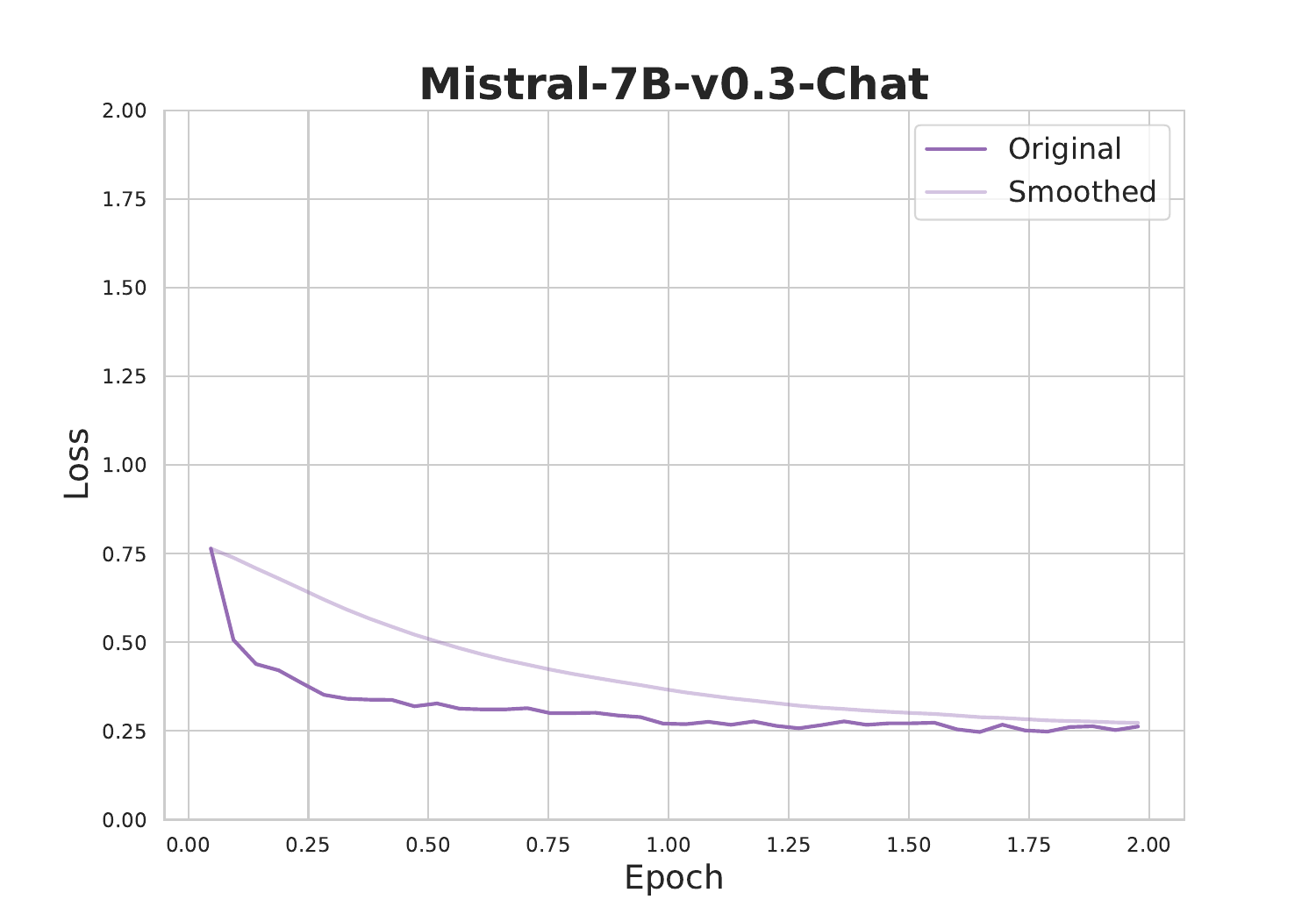}
  \caption{\small{Fine-tuning loss plots for Mistral-7B-v0.3-Chat.}}
\end{figure}

\begin{table}[h]
    \centering
    \resizebox{1\columnwidth}{!}{   
    \begin{tabular}{ll}
        \toprule
        \textbf{Parameter} & \textbf{Value} \\ \midrule
        Base Model & mistralai/Mistral-7B-Instruct-v0.3 \\ 
        ROUGE-L & 0.694 \\ 
        ROUGE-1 & 0.719 \\ 
        Fine-Tuning Type & LoRA \\ 
        LoRA Alpha & 16 \\ 
        LoRA Rank & 8 \\ 
        Cutoff Length & 4096 \\ 

        Gradient Accumulation Steps & 8 \\ 
        Learning Rate & 5.0e-05 \\ 
        LR Scheduler Type & Cosine \\ 
        Number of Training Epochs & 2.0 \\ 
        Optimizer & AdamW \\ 
        
        Quantization Bit & 4 \\ 
        \bottomrule
    \end{tabular}
    }
    \caption{Fine-tuning Parameters for Mistral-7B-v0.3-Chat}
    \label{tab:training_params_Mistral-7B-v0.3-Chat}
\end{table}

\begin{figure}[h]
\centering
  \includegraphics[width=1\linewidth]{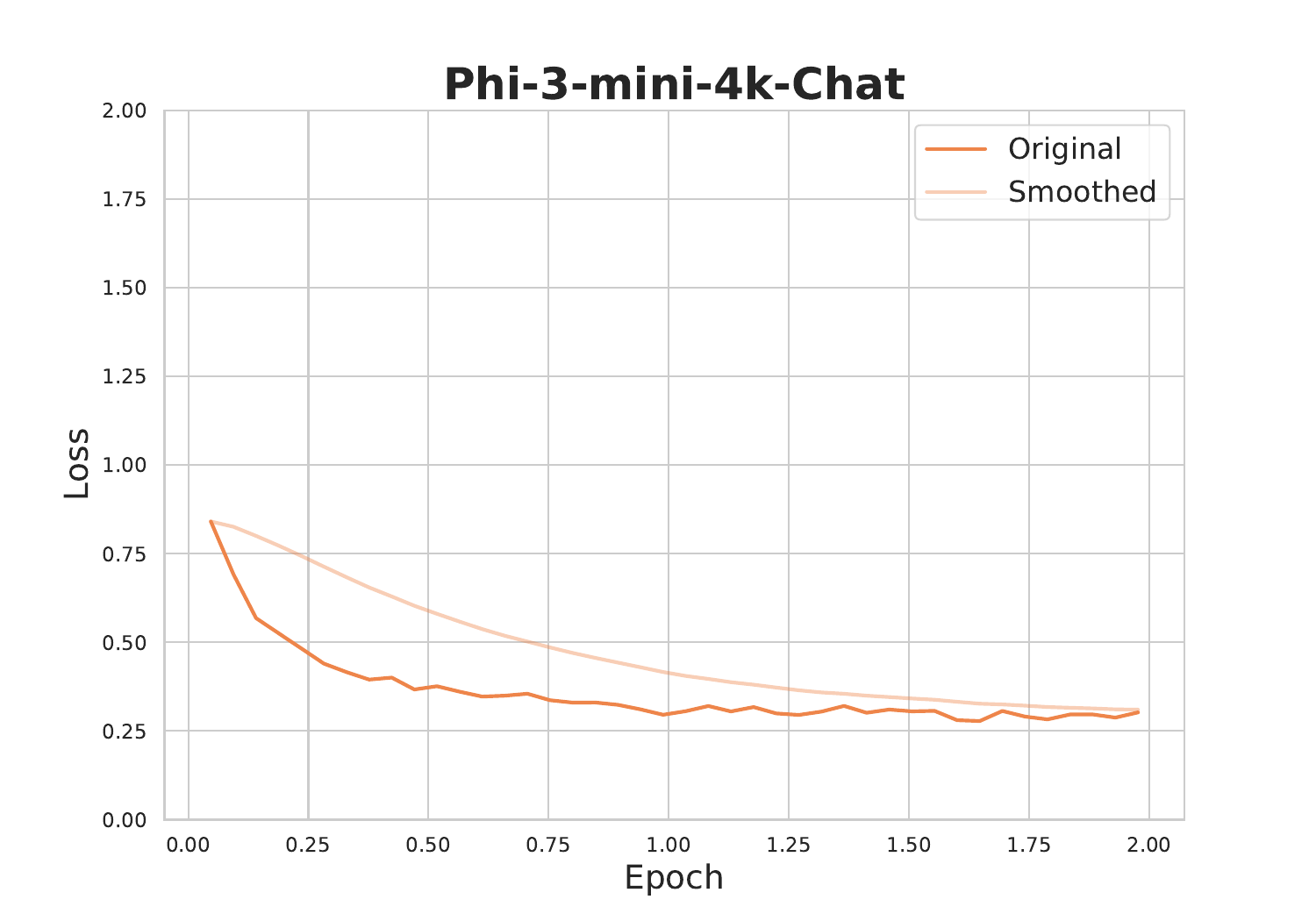}
  \caption{\small{Fine-tuning loss plots for Phi-3-mini-4k-Chat.}}
\end{figure}

\begin{table}[h]
    \centering
    \resizebox{1\columnwidth}{!}{   
    \begin{tabular}{ll}
        \toprule
        \textbf{Parameter} & \textbf{Value} \\ \midrule
        Base Model & microsoft/Phi-3-mini-4k-instruct \\ 
        ROUGE-L & 0.658 \\ 
        ROUGE-1 & 0.685 \\ 
        Fine-Tuning Type & LoRA \\ 
        LoRA Alpha & 16 \\ 
        LoRA Rank & 8 \\ 
        Cutoff Length & 4096 \\ 

        Gradient Accumulation Steps & 8 \\ 
        Learning Rate & 5.0e-05 \\ 
        LR Scheduler Type & Cosine \\ 
        Number of Training Epochs & 2.0 \\ 
        Optimizer & AdamW \\ 
        
        Quantization Bit & 4 \\ 
        \bottomrule
    \end{tabular}
    }
    \caption{Fine-tuning Parameters for Phi-3-mini-4k-Chat}
    \label{tab:training_params_Phi-3-mini-4k-Chat}
\end{table}

\begin{figure}[h]
\centering
  \includegraphics[width=1\linewidth]{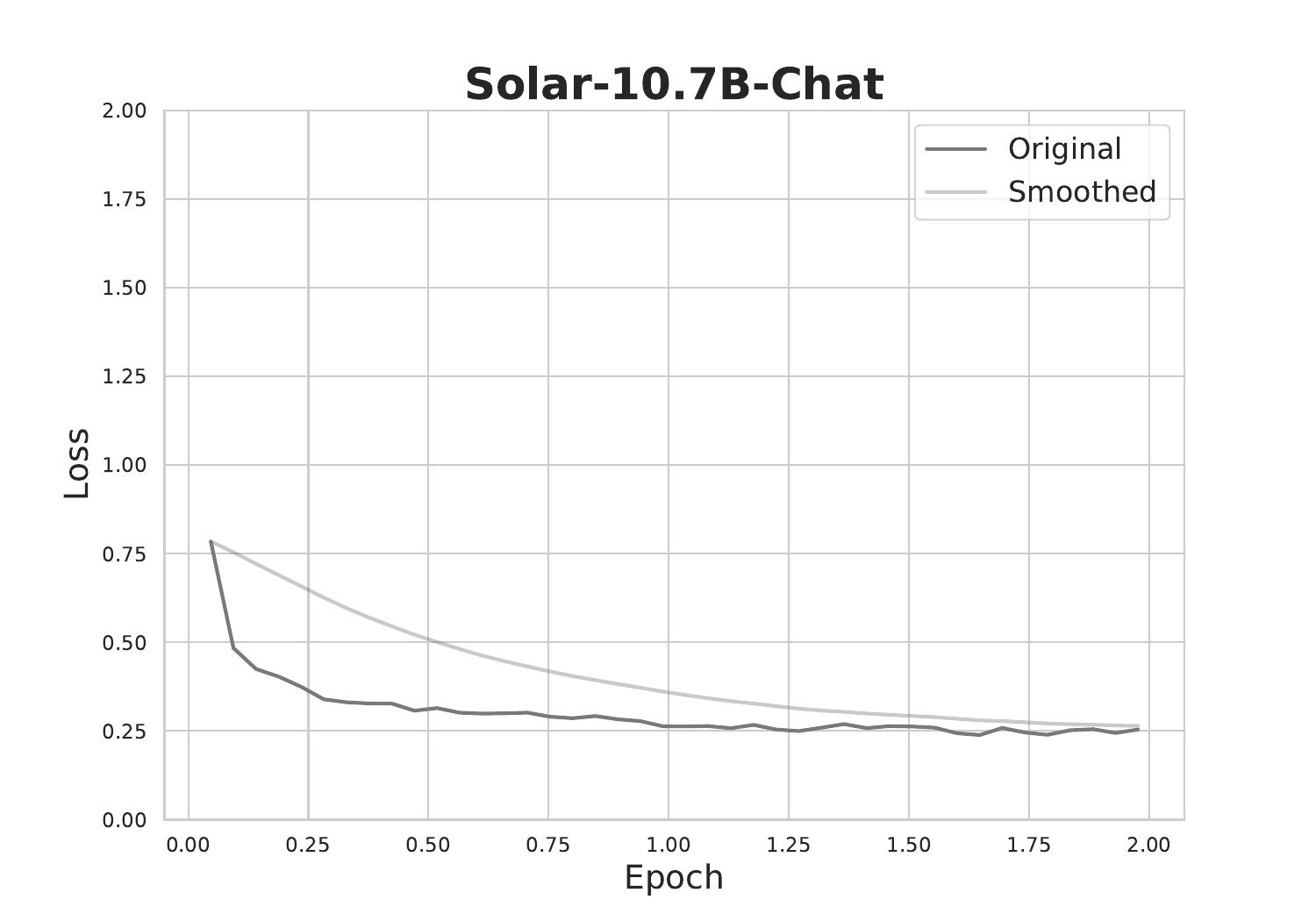}
  \caption{\small{Fine-tuning loss plots for SOLAR-10.7B-Chat.}}
\end{figure}

\begin{table}[h]
    \centering
    \resizebox{1\columnwidth}{!}{   
    \begin{tabular}{ll}
        \toprule
        \textbf{Parameter} & \textbf{Value} \\ \midrule
        Base Model & upstage/SOLAR-10.7B-Instruct-v1.0 \\ 
        ROUGE-L & 0.689 \\ 
        ROUGE-1 & 0.714 \\ 
        Fine-Tuning Type & LoRA \\ 
        LoRA Alpha & 16 \\ 
        LoRA Rank & 8 \\ 
        Cutoff Length & 4096 \\ 

        Gradient Accumulation Steps & 8 \\ 
        Learning Rate & 5.0e-05 \\ 
        LR Scheduler Type & Cosine \\ 
        Number of Training Epochs & 2.0 \\ 
        Optimizer & AdamW \\ 
        
        Quantization Bit & 4 \\ 
        \bottomrule
    \end{tabular}
    }
    \caption{Fine-tuning Parameters for SOLAR-10.7B-Chat}
    \label{tab:training_params_SOLAR-10.7B-Chat}
\end{table}

\begin{figure}[h]
\centering
  \includegraphics[width=1\linewidth]{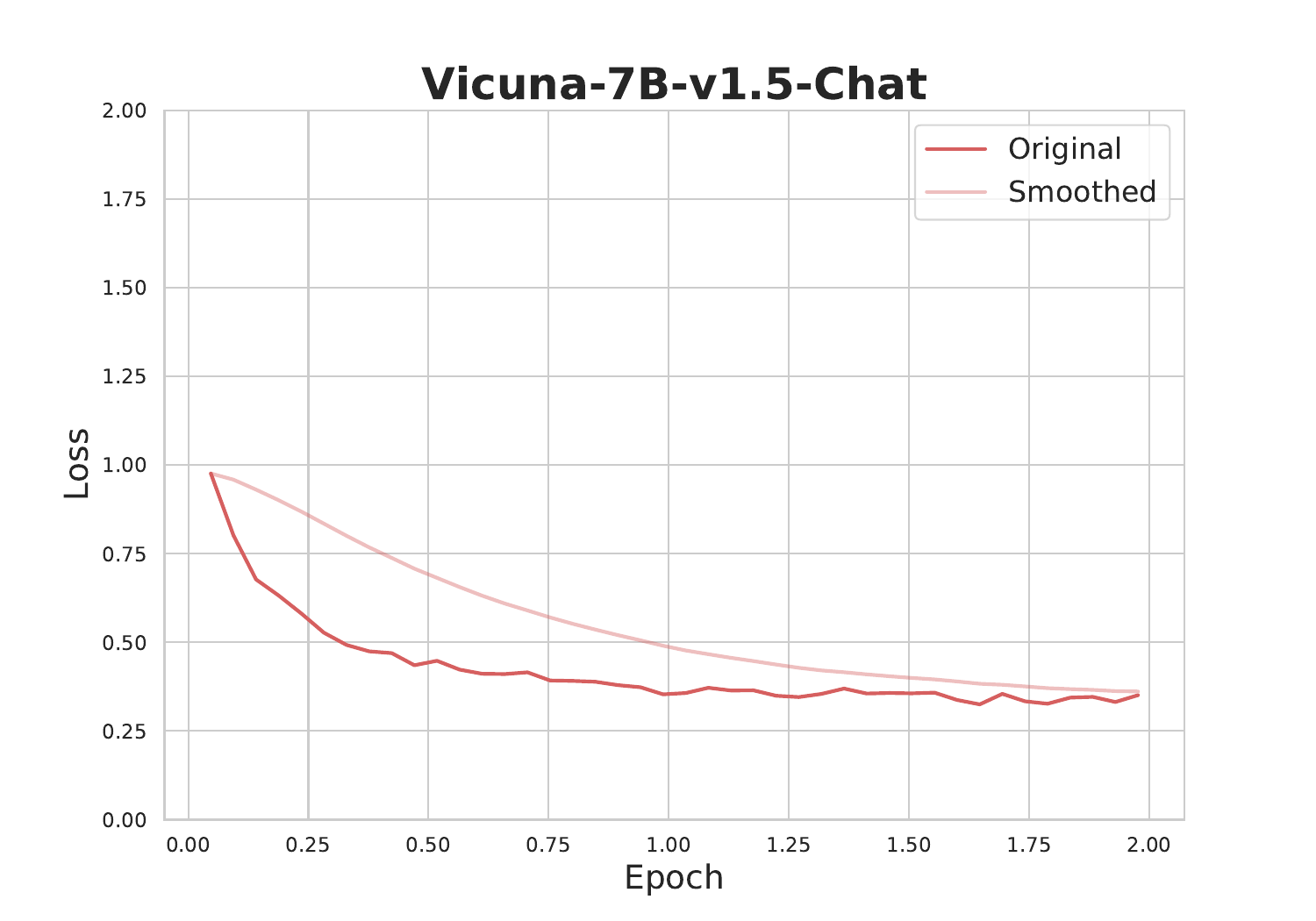}
  \caption{\small{Fine-tuning loss plots for Vicuna-7B-v1.5-Chat.}}
\end{figure}

\begin{table}[h]
    \centering
    \resizebox{1\columnwidth}{!}{   
    \begin{tabular}{ll}
        \toprule
        \textbf{Parameter} & \textbf{Value} \\ \midrule
        Base Model & lmsys/vicuna-7b-v1.5 \\ 
        ROUGE-L & 0.676 \\ 
        ROUGE-1 & 0.700 \\ 
        Fine-Tuning Type & LoRA \\ 
        LoRA Alpha & 16 \\ 
        LoRA Rank & 8 \\ 
        Cutoff Length & 4096 \\ 

        Gradient Accumulation Steps & 8 \\ 
        Learning Rate & 5.0e-05 \\ 
        LR Scheduler Type & Cosine \\ 
        Number of Training Epochs & 2.0 \\ 
        Optimizer & AdamW \\ 
        
        Quantization Bit & 4 \\ 
        \bottomrule
    \end{tabular}
    }
    \caption{Fine-tuning Parameters for Vicuna-7B-v1.5-Chat}
    \label{tab:training_params_Vicuna-7B-v1.5-Chat}
\end{table}

\begin{figure}[h]
\centering
  \includegraphics[width=1\linewidth]{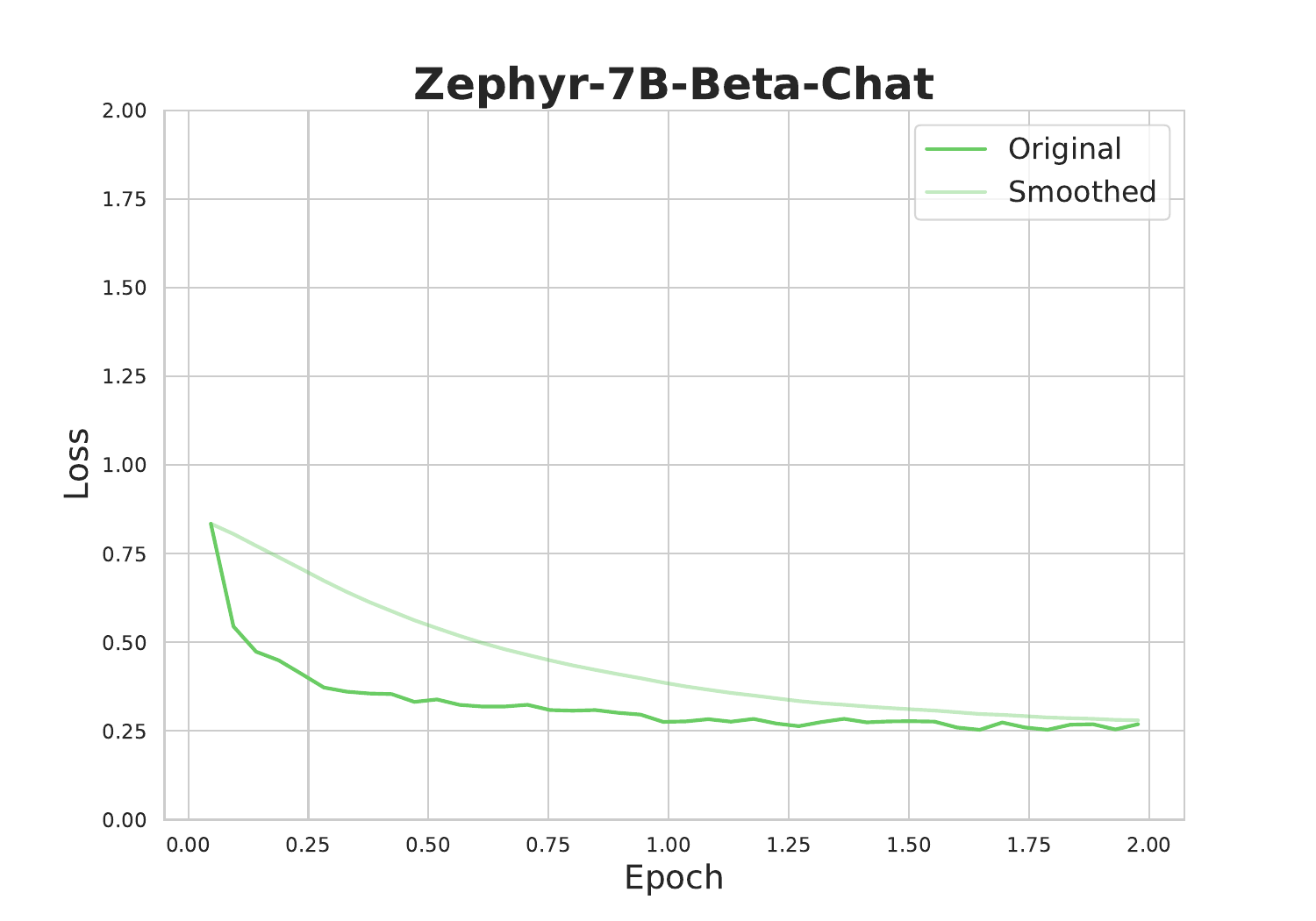}
  \caption{\small{Fine-tuning loss plots for Zephyr-7B-Beta-Chat.}}
\end{figure}

\begin{table}[h]
    \centering
    \resizebox{1\columnwidth}{!}{   
    \begin{tabular}{ll}
        \toprule
        \textbf{Parameter} & \textbf{Value} \\ \midrule
        Base Model & HuggingFaceH4/zephyr-7b-beta \\ 
        ROUGE-L & 0.686 \\ 
        ROUGE-1 & 0.712 \\ 
        Fine-Tuning Type & LoRA \\ 
        LoRA Alpha & 16 \\ 
        LoRA Rank & 8 \\ 
        Cutoff Length & 4096 \\ 

        Gradient Accumulation Steps & 8 \\ 
        Learning Rate & 5.0e-05 \\ 
        LR Scheduler Type & Cosine \\ 
        Number of Training Epochs & 2.0 \\ 
        Optimizer & AdamW \\ 
        
        Quantization Bit & 4 \\ 
        \bottomrule
    \end{tabular}
    }
    \caption{Fine-tuning Parameters for Zephyr-7B-Beta-Chat}
    \label{tab:training_params_Zephyr-7B-Beta-Chat}
\end{table}

\begin{figure}[h]
\centering
  \includegraphics[width=1\linewidth]{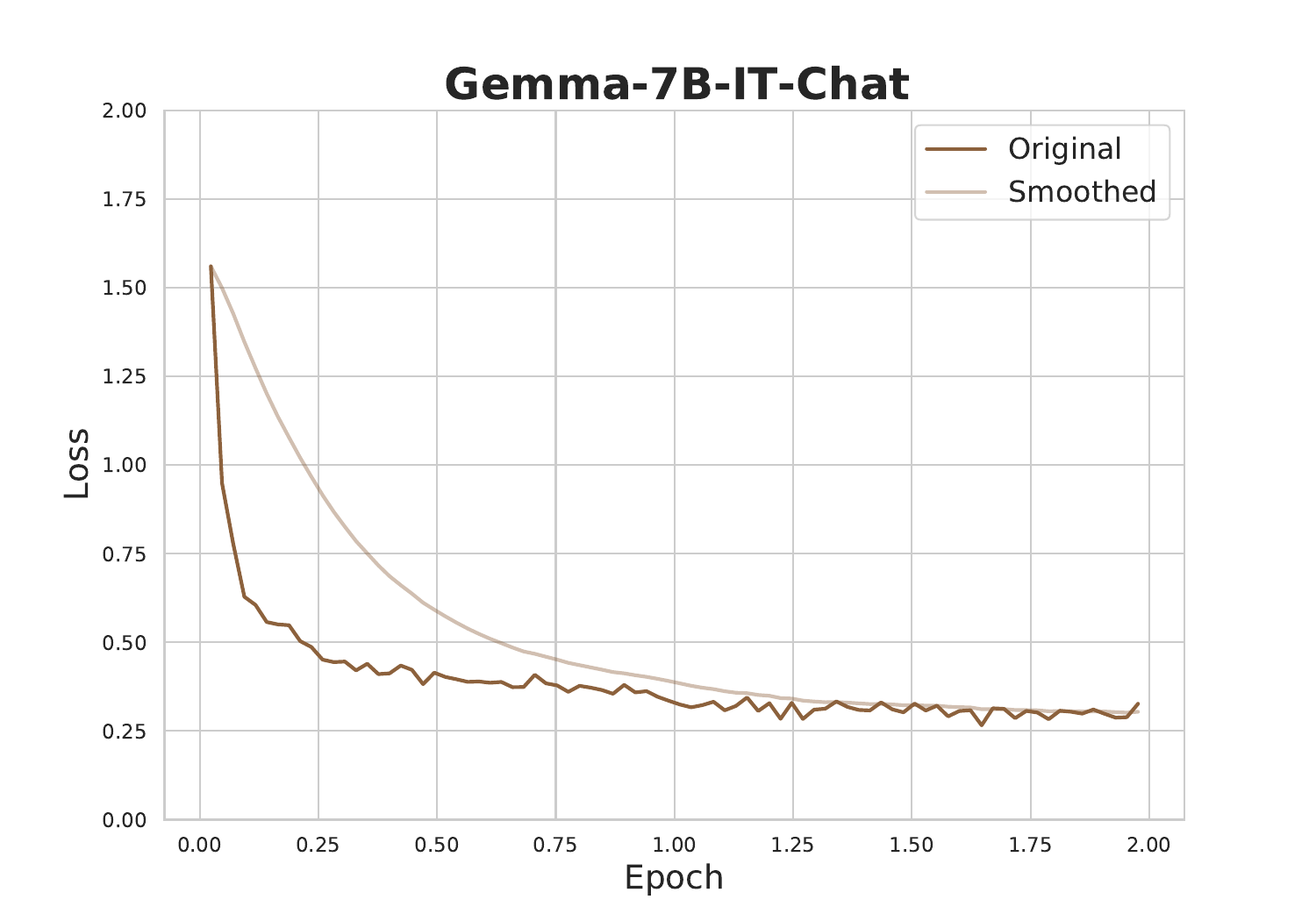}
  \caption{\small{Fine-tuning loss plots for Gemma-7B-IT-Chat.}}
\end{figure}

\begin{table}[h]
    \centering
    \resizebox{1\columnwidth}{!}{   
    \begin{tabular}{ll}
        \toprule
        \textbf{Parameter} & \textbf{Value} \\ \midrule
        Base Model & google/gemma-7b-it \\ 
        ROUGE-L & 0.678 \\ 
        ROUGE-1 & 0.703 \\ 
        Fine-Tuning Type & LoRA \\ 
        LoRA Alpha & 16 \\ 
        LoRA Rank & 8 \\ 
        Cutoff Length & 4096 \\ 

        Gradient Accumulation Steps & 8 \\ 
        Learning Rate & 5.0e-05 \\ 
        LR Scheduler Type & Cosine \\ 
        Number of Training Epochs & 2.0 \\ 
        Optimizer & AdamW \\ 
        
        Quantization Bit & 4 \\ 
        \bottomrule
    \end{tabular}
    }
    \caption{Fine-tuning Parameters for Gemma-7B-IT-Chat}
    \label{tab:training_params_Gemma-7B-IT-Chat}
\end{table}

\begin{figure}[h]
\centering
  \includegraphics[width=1\linewidth]{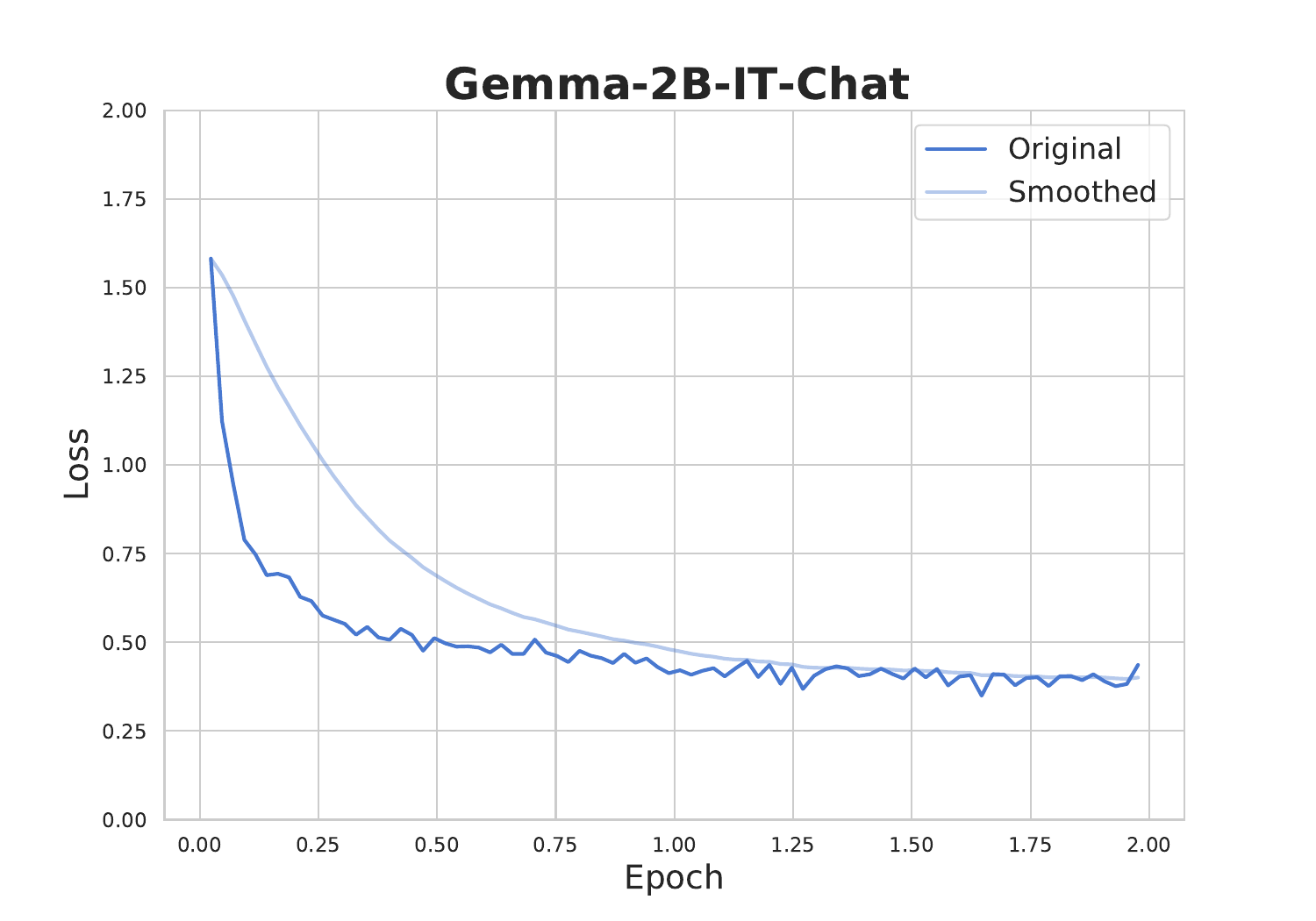}
  \caption{\small{Fine-tuning loss plots for Gemma-2B-IT-Chat.}}
\end{figure}

\begin{table}[h]
    \centering
    \resizebox{1\columnwidth}{!}{   
    \begin{tabular}{ll}
        \toprule
        \textbf{Parameter} & \textbf{Value} \\ \midrule
        Base Model & google/gemma-2b-it \\ 
        ROUGE-L & 0.667 \\ 
        ROUGE-1 & 0.692 \\ 
        Fine-Tuning Type & LoRA \\ 
        LoRA Alpha & 16 \\ 
        LoRA Rank & 8 \\ 
        Cutoff Length & 4096 \\ 

        Gradient Accumulation Steps & 8 \\ 
        Learning Rate & 5.0e-05 \\ 
        LR Scheduler Type & Cosine \\ 
        Number of Training Epochs & 2.0 \\ 
        Optimizer & AdamW \\ 
        Quantization Bit & 4 \\ 
        \bottomrule
    \end{tabular}
    }
    \caption{Fine-tuning Parameters for Gemma-2B-IT-Chat}
    \label{tab:training_params_Gemma-2B-IT-Chat}
\end{table}

% \begin{figure}[t] % Use [ht] for "here" or "top" placement of the float
% \centering % Centers the minipage on the page
% \fbox{%
% \begin{minipage}{0.99\columnwidth} % Makes the minipage the width of the column
% \scriptsize % Sets a smaller font size
% \texttt{%
% Analyze text against provided triplets, classifying claims as "Attributable", "Contradictory", or "Extrapolatory".\newline
% Justify your classification using the following structure:\newline
% - "text\_span": Text under evaluation.\newline
% - "prediction": Category of the text (Attributable/Contradictory/Extrapolatory).\newline
% - "triplets": Relevant triplets (if any, else "NA").\newline
% - "rationale": Reason for classification.\newline
% For multiple claims, number each component (e.g., "text\_span1", "prediction1",..). Use "NA" for inapplicable keys.\newline
% Example:\newline
% "text\_span1": "Specific claim",\newline
% "prediction1": "Attributable/Contradictory/Extrapolatory",\newline
% "triplets1": "Relevant triplets",\newline
% "rationale1": "Prediction justification",\newline
% ...\newline
% Input for analysis:\newline
% -Text: \{Input Text\}\newline
% -Triplets: \{Retrieved Triplets\}\newline
% }
% \vspace{-2mm}
% \end{minipage}
% }
% \caption{Prompt 1 GPT.}
% \label{fig:prompt1}
% \end{figure}

\clearpage
% \newpage
% \subsection{Prompt Details}
% \label{app:prompt}

\begin{figure*}[t] % Use [ht] for "here" or "top" placement of the float
\centering % Centers the minipage on the page
\resizebox{0.85\textwidth}{!}{  
\fbox{%
\begin{minipage}{0.99\textwidth} % Makes the minipage the width of the column
\scriptsize % Sets a smaller font size
\texttt{%
**Text Span Attribution Verification**\newline
\newline
**Objective:** Predict whether the text span is "Attributable", "Contradictory", or "Extrapolatory" based on the information provided in the triplets.\newline
\newline
**Instructions:**\newline
\newline
1. **Read the Full Text:**\newline
- Understand the context and content of the full text string.\newline
\newline
2. **Examine the Text Span:**\newline
- Determine the claims made within the text span.\newline
\newline
3. **Analyze the Triplets:**\newline
- Evaluate if the triplets support, refute, or neither support nor refute the claims in the text span.\newline
\newline
4. **Make Your Prediction:**\newline
- Classify the text span as "Attributable", "Contradictory", or "Extrapolatory" based on your analysis of the triplets.\newline
\newline
5. **Provide Rationale:**\newline
- Clearly explain your reasoning for the classification.\newline
\newline
**Classification Criteria:**\newline
\newline
- **"Attributable"**: The text span is sufficiently supported by the triplet(s). All claims in the text span are directly present in the triplet information.\newline
- **"Contradictory"**: The text span is conclusively refuted by the triplet(s). All claims in the text span are directly contradicted by the triplet information.\newline
- **"Extrapolatory"**: The triplet(s) can neither support nor refute the text span. The information provided is either irrelevant, indirect, or related but not sufficient to support or refute the text span.\newline
\newline
**Example:**\newline
\newline
**Full Text:** "Albert Einstein is widely recognized as the father of modern physics. He was awarded the Nobel Prize in Physics for his services to Theoretical Physics."\newline
\newline
**Text Span:** "He was awarded the Nobel Prize in Physics."\newline
\newline
**Triplets:** [("Albert Einstein", "award received", "Nobel Prize in Physics")]\newline
\newline
**Sample Evaluation:**\newline
- **Prediction:** "Attributable"\newline
- **Rationale:** "The triplet directly supports the claim that Albert Einstein received the Nobel Prize in Physics."\newline
\newline
**Example:**\newline
\newline
**Full Text:** "Isaac Newton discovered the element radium."\newline
\newline
**Text Span:** "Isaac Newton discovered radium."\newline
\newline
**Triplets:** [("Marie Curie", "discovered", "radium")]\newline
\newline
**Sample Evaluation:**\newline
- **Prediction:** "Contradictory"\newline
- **Rationale:** "The triplet states that Marie Curie discovered radium, contradicting the claim that Isaac Newton discovered it."\newline
\newline
**Example:**\newline
\newline
**Full Text:** "The Eiffel Tower is a wrought-iron lattice tower that was opened in 1889."\newline
\newline
**Text Span:** "The Eiffel Tower is a wrought-iron lattice tower that was opened in 1889."\newline
\newline
**Triplets:** [("Eiffel Tower", "located in", "Paris")]\newline
\newline
**Sample Evaluation:**\newline
- **Prediction:** "Extrapolatory"\newline
- **Rationale:** "The triplet states that the Eiffel Tower is located in Paris, which is related but not sufficient to confirm or refute that it was opened in 1889."\newline
\newline
**Verification Checklist:**\newline
\newline
- [ ] The prediction accurately reflects the relationship between the text span and the triplets.\newline
- [ ] The rationale clearly explains the classification based on the triplets.\newline
- [ ] The explanation is free from irrelevant information.\newline
\newline
**Response Format:**\newline
Provide your evaluation in the following JSON format:\newline
- "prediction": "Attributable", "Contradictory", or "Extrapolatory"\newline
- "rationale": "Your comments here"\newline
\newline
**Inputs to Evaluate**\newline
\newline
**Full text:** "\{full\_text\}"\newline
**Text span:** "\{text\_span\}"\newline
**Triplets:** \{triplets\}\newline
}
% \vspace{-2mm}
\end{minipage}
}
}
\caption{Prompt for the text span attribution verification task, guiding the model to classify text spans as "Attributable," "Contradictory," or "Extrapolatory" based on the provided triplets. The prompt design incorporates concepts such as few-shot learning, chain-of-thought reasoning \cite{kojima2022large}, and tailored prompt engineering \cite{openai_promptengineering,nori2023can}}
\label{fig:App_prompt2}
\end{figure*}

\end{document}